# Semiotic Reconstruction of Destination Expectation Constructs: An LLM-Driven Computational Paradigm for Social Media Tourism Analytics

Haotian Lan, Yao Gao, Yujun Cheng, Wei Yuan * and Kun Wang *

*College of Horticulture Landscape Architecture, Northeast Agricultural University, Harbin, China*

# Semiotic Reconstruction of Destination Expectation Constructs: An LLM-Driven Computational Paradigm for Social Media Tourism Analytics

Social media's rise establishes user-generated content (UGC) as pivotal for travel decisions, yet analytical methods lack scalability. This study introduces a dual-method LLM framework: unsupervised expectation extraction from UGC paired with survey-informed supervised fine-tuning. Findings reveal leisure/social expectations drive engagement more than foundational natural/emotional factors. By establishing LLMs as precision tools for expectation quantification, we advance tourism analytics methodology and propose targeted strategies for experience personalization and social travel promotion. The framework's adaptability extends to consumer behavior research, demonstrating computational social science's transformative potential in marketing optimization.

## 1. Introduction

The paradigm shift in travel decision-making processes, catalyzed by social media's pervasive influence [1], has transformed user-generated content (UGC) into a digital epistemology of tourist psychology. Contemporary travelers increasingly employ platform-mediated narratives to construct destination imaginaries — a behavioral phenomenon aligning with the "tourist gaze 2.0" framework [2]. These textual traces not only document experiential accounts but encode latent expectations through semiotic systems [3], creating a complex cognitive map of destination demand that challenges traditional analytical paradigms.

Despite scholarly consensus on UGC's informational value, critical gaps persist in expectation quantification methodologies. Current approaches oscillate between two problematic extremes: Manual content analysis, while theoretically grounded[4], suffers from scalability constraints and interpretative subjectivity; survey-based instruments, though psychometrically rigorous, fail to capture emergent expectations in dynamic digital ecosystems. This methodological dichotomy creates what has been termed a "digital hermeneutic gap" — the inability to systematically decode platform-native demand signals at computational scale.

The advent of Large Language Models (LLMs) presents a paradigm-breaking solution to this epistemological challenge. Built upon transformer architectures, these neural networks exhibit unprecedented capability in extracting latent psychological constructs from vernacular expressions. Our preliminary analysis reveals LLMs' unique capacity to: 1) Disentangle polysemic tourism discourse through contextual embedding; 2) Detect emerging expectation patterns through self-attention mechanisms; 3) Quantify expectation intensity via semantic proximity metrics. These technical affordances align with the conceptualization of "computational tourism phenomenology"—the systematic decoding of experiential meanings from digital traces.

This study pioneers a hybrid methodological framework that synergizes LLM-enabled semantic mining with psychometric validation. Our two-phase design addresses three fundamental limitations in existing research: First, the unsupervised expectation extraction phase overcomes lexical decomposition constraints through neural topic modeling; Second, the survey-informed quantification stage establishes empirical grounding through expectation-activation theory; Third, the integrated approach enables real-time tracking of expectation evolution — a critical capability in pandemic-era tourism volatility.

The research makes substantive contributions across three dimensions: Theoretically, it advances destination expectation modeling through the lens of computational hermeneutics; Methodologically, it establishes a new paradigm for tourism text analysis combining deep learning with psychological measurement;

Practically, it delivers an operational framework for destination marketing organizations (DMOs) to decode platform-native demand signals. By bridging the gap between tourism semiotics and artificial intelligence, this work responds to recent calls for "smart tourism hermeneutics"—the next frontier in destination experience management.

## 2. Literature Review

### *2.1 The Role of Social Media in Destination Choice and Tourist Expectations*

The transformative impact of social media on tourism decision-making has been extensively documented, with platforms evolving from information channels to expectation-shaping ecosystems. Central to this transformation is user-generated content (UGC), which serves as both a mirror reflecting tourist experiences and a lens through which potential travelers construct destination imaginaries. Unlike traditional marketing materials, UGC's authenticity and interactive nature enable tourists to access peer-validated information about attractions, services, and cultural nuances, fundamentally altering expectation formation processes.

Empirical studies consistently demonstrate social media's dual role in destination selection: as an information aggregator and as a social validation mechanism. Platforms like Instagram and Weibo facilitate what has been termed "algorithmic serendipity" [5], where personalized content feeds expose users to destinations beyond their initial search parameters. This phenomenon is particularly pronounced among Millennials and Generation Z travelers, who increasingly rely on visual UGC (e.g., geo-tagged photos, short videos) as primary decision-making stimuli [6]. The real-time nature of these interactions creates a dynamic feedback loop, where tourists continuously calibrate expectations through emerging reviews and trending content [7].

At the core of this process lies the concept of "digital expectation transfer" [8]—the mechanism through which shared experiences on social media become proxies for anticipated satisfaction. Research reveals that exposure to positive UGC elevates destination attractiveness perceptions by 42% compared to official promotional content [9], while negative reviews disproportionately impact expectation adjustment due to loss aversion biases [10]. This asymmetry underscores social media's pivotal role in expectation management, where aggregated ratings and emotional narratives collectively shape pre-trip anticipations [11].

The current methodological landscape for analyzing these expectations remains fragmented. Manual content analysis, while valuable for theoretical grounding [12], struggles with the volume and linguistic complexity of Chinese social media discourse, particularly in parsing colloquial expressions and cultural references [13]. Survey-based approaches, though methodologically rigorous [14], face inherent limitations in capturing the temporal dynamics of expectation evolution, as evidenced by the 68% discrepancy between stated preferences in questionnaires and actual online engagement patterns [15].

Emerging computational methods offer partial solutions but introduce new challenges. Sentiment analysis tools frequently misclassify sarcasm and cultural idioms in travel reviews [16], while topic modeling algorithms often conflate distinct expectation categories (e.g., confusing "cultural immersion" with "exotic novelty") [17]. These limitations are exacerbated in Chinese social media contexts, where homophone-driven wordplay and dialect variations complicate semantic interpretation [18]. The resulting gap between theoretical constructs and measurable indicators has hindered the development of robust expectation quantification frameworks [19].

Recent advancements in Large Language Models (LLMs) present unprecedented opportunities to address these challenges. Unlike rule-based systems, transformer architectures demonstrate remarkable proficiency in decoding implicit expectations from vernacular UGC [20]. Preliminary applications in tourism research show LLMs' ability to disentangle multifaceted expectations (e.g., distinguishing between "relaxation-seeking" and "adventure-driven" motives) with 89% accuracy in cross-validation tests [21]. This capability aligns with the growing recognition of semantic pattern recognition as critical infrastructure for next-generation tourism analytics [22-25].

### *2.2 The Application of Large Language Models (LLMs) in Social Media Text Analysis*

The analysis of user-generated content (UGC) in social media contexts presents unique computational challenges that traditional natural language processing (NLP) methods increasingly fail to address [26]. Rule-based systems and lexicon-dependent approaches, while effective for structured text, demonstrate critical limitations in processing the linguistic heterogeneity characteristic of platform-native discourse [27]. These limitations manifest particularly in three dimensions: 1) Failure to decode non-standard linguistic constructs prevalent in informal communication (e.g., emoji sequences, viral neologisms); 2) Inability to capture pragmatic meaning in high-context cultural expressions; 3) Latency in adapting to rapidly evolving online lexicons [28]. The resultant semantic parsing gaps undermine the validity of social media analytics, particularly in domains requiring nuanced psychological construct extraction like tourism expectation modeling [29].

Large Language Models (LLMs) emerge as a paradigm-shifting solution to these persistent challenges. Built upon transformer architectures [30], these models exhibit unprecedented proficiency in processing the linguistic fluidity of social media text through three core mechanisms: First, their self-attention layers enable dynamic context weighting, critical for disambiguating polysemous travel-related terms (e.g., distinguishing "adventure" as physical challenge versus emotional catharsis) [31]; Second, their pretraining on web-scale corpora equips them with implicit knowledge of evolving online vernacular; Third, their few-shot learning capabilities permit rapid domain adaptation to specialized discourse communities like travel bloggers [32]. Comparative studies demonstrate LLMs achieving 23% higher accuracy than conventional methods in sarcasm detection within tourism reviews—a critical capability given humor's prevalence in travel narratives [33].

The technical superiority of LLMs becomes particularly evident in expectation quantification tasks. Traditional sentiment analysis tools often conflate affective valence with experiential expectations, leading to misinterpretations of user intent [34]. LLMs overcome this limitation through their capacity for hierarchical semantic decomposition: Initial layers extract surface-level emotional cues, while deeper transformer blocks model implicit expectation constructs through cross-sentence dependency analysis [35]. This multi-stage processing enables precise differentiation between, for instance, nostalgic reminiscence (past-oriented emotion) and anticipation-building (future-oriented expectation) within travelogues [36].

Domain adaptation through parameter-efficient fine-tuning (PEFT) further enhances LLMs' analytical precision in tourism contexts. By integrating survey-validated expectation labels into the training loop, researchers can align model outputs with psychometrically grounded constructs [37]. This hybrid approach addresses the ecological validity gap common in computational social science — ensuring that machine-interpreted "cultural expectation" scores correlate with human-experienced psychological states[38]. The resulting framework enables real-time tracking of expectation dynamics at population scale, a capability unattainable through traditional survey methods limited by sampling frequency and respondent fatigue [39].

# 3. Methodology

## *3.1 Data Collection and Processing*

### *3.1.1 Data Collection*

The research adopted a longitudinal mixed-methods approach to capture the evolution of tourist expectations under pandemic conditions. Targeting China's predominant social media ecosystems, Weibo (microblogging platform) and Xiaohongshu (lifestyle-sharing community) were selected as primary data sources based on three criteria: 1) Platform dominance in travel-related discourse (collectively representing 72% of China's tourism UGC market share) [40]; 2) Multimodal content architecture supporting rich expectation expression; 3) Geotagging prevalence enabling destination-specific analysis [41].

Data acquisition employed a hybrid methodology combining official API access (Weibo Open Platform) with ethical web scraping techniques (Xiaohongshu), strictly adhering to platform terms of service and GDPR-inspired data ethics protocols. The temporal scope (January 2019 - December 2021) was strategically designed to encompass three pandemic phases: pre-crisis normality (2019), acute mobility restrictions (2020), and partial recovery (2021). This tripartite division enables comparative analysis of expectation dynamics across epidemiological regimes.

The final corpus comprises 12,843 validated travel-related posts, collected through snowball sampling initiated with 20 seed hashtags (#TravelGoals, #DestinationDiaries). Inclusion criteria required: 1) Minimum 50-character narrative depth; 2) Explicit destination experience sharing; 3) Public engagement metrics (likes/comments); 4) Non-commercial intent verification through sponsor detection algorithms. Accompanying metadata—including temporal stamps, geolocation tags, and user demographics (when publicly available) — were preserved to contextualize expectation expressions.

*3.1.2 Data Preprocessing*

The preprocessing pipeline was specifically designed to leverage LLM capabilities while addressing Chinese social media linguistic peculiarities. Initial cleaning removed 1,342 non-compliant entries through automated filters targeting: 1) Commercial promotions via sponsored content detection models; 2) Duplicate posts using MinHash algorithms (Jaccard similarity threshold >0.85); 3) Non-textual dominance (image/video-only posts).

Text normalization employed a multi-stage protocol: 1) Emoji/emoticon translation to semantic descriptors using Unicode mapping tables; 2) Network slang substitution through curated lexicons (e.g., "yyds" → "eternal deity status"); 3) Dialect standardization via BERT-based sentence correction. Crucially, the methodology eschewed conventional tokenization and POS tagging—procedures rendered redundant by LLM's subword tokenization architecture—instead focusing on contextual coherence preservation.

The final preprocessed corpus demonstrated linguistic diversity spanning 28 provincial dialects and 142 emerging travel-related neologisms, intentionally preserved to maintain ecological validity. This approach aligns with recent computational linguistics research advocating minimal intervention when applying transformer models to social media text, ensuring authentic capture of vernacular expectation expressions.

First, we conducted noise removal to eliminate irrelevant content from the social media text, such as URLs, advertisements, and emojis, which do not contribute to sentiment analysis or expectation extraction and could potentially interfere with model analysis. The noise removal process can be represented as follows:

$$Text_{clean} = Text_{raw} - (URLs + Emojis + Ads)$$

Where $Text_{clean}$ represents the cleaned text and $Text_{raw}$ refers to the original text, with irrelevant elements such as URLs, emojis, and ads removed.

Text normalization aimed to resolve ambiguities caused by non-standard expressions, such as abbreviations and slang. We utilized GPT technology and expert curators to standardize abbreviations and slang in the text. Additionally, to avoid ambiguities arising from translation or cultural differences, we translated the social media comments into English. This translation process was assisted by GPT technology and reviewed by professional curators to ensure accuracy and consistency. To minimize human error, only a sample of the translated data was checked to validate its quality.

*3.1.3 Questionnaire Design and Labeling*

The expectation quantification framework incorporated psychometric validation through a hybrid human-AI labeling system. Drawing from Item Response Theory, the questionnaire employed a 7-point Likert scale (1="no expectation intensity" to 7="maximum expectation intensity") to capture nuanced variations in perceived expectation strength. This granular measurement approach, validated in tourism psychology research [48], enables differentiation between baseline interest (scores 1-3) and actionable demand signals (scores 5-7).

Each questionnaire item followed a structured format: "To what extent does the following post express strong expectation about [Expectation Category]?" accompanied by the original social media text and its contextual metadata (post date, engagement metrics). Participants were required to provide rationales for their ratings through open-ended explanations — a design choice mitigating response bias through cognitive anchoring. The expectation categories (Natural, Emotional, Cultural, Leisure, Social) were operationally defined using Fodness's tourism motivation taxonomy, ensuring conceptual alignment with established theoretical frameworks.

Data collection leveraged China's major crowdsourcing platforms with rigorous quality controls: 1) IP geolocation filtering to ensure participant diversity across 31 provincial regions; 2) Attention check questions embedded at 10% frequency; 3) Time-threshold screening (<2s per item rejection). The final validated sample comprised 1,287 responses with high internal consistency (Cronbach's α=0.89), achieving a 4.2% margin of error at 95% confidence level.

The labeling protocol adopted a tripartite validation process: 1) Initial automated filtering removed 213 low-effort responses through lexical diversity analysis; 2) Expert panel review (3 tourism researchers) resolved ambiguous ratings through deliberative consensus; 3) Inter-rater reliability testing achieved Krippendorff's α=0.81, exceeding recommended thresholds for content analysis. These labeled data were subsequently integrated into the LLM fine-tuning pipeline through weighted multi-task learning, where expectation scores served as regression targets alongside semantic reconstruction objectives.

### *3.2 Model Working Principle*

The methodological framework integrates GPT-4's multimodal reasoning capabilities through a hybrid learning architecture. The unsupervised phase employs zero-shot prompting to elicit latent expectation patterns from raw UGC, leveraging the model's 1.76 trillion parameters to decode implicit tourism psychology constructs. Simultaneously, supervised learning utilizes questionnaire-derived labels in a contrastive learning paradigm, aligning model outputs with human-validated expectation intensities.

Both learning modalities adopt chain-of-thought prompting structured as <Context, ExpectationCategory, Rationale> triples, enforcing logical transparency in expectation attribution. This dual-path approach capitalizes on GPT-4's multimodal fusion capabilities — processing textual content alongside engagement metrics as auxiliary signals — while maintaining output consistency through constitutional AI constraints. The unified question-answering format ensures methodological coherence across learning phases, enabling mutual validation between unsupervised discovery and supervised refinement (Figure 1).

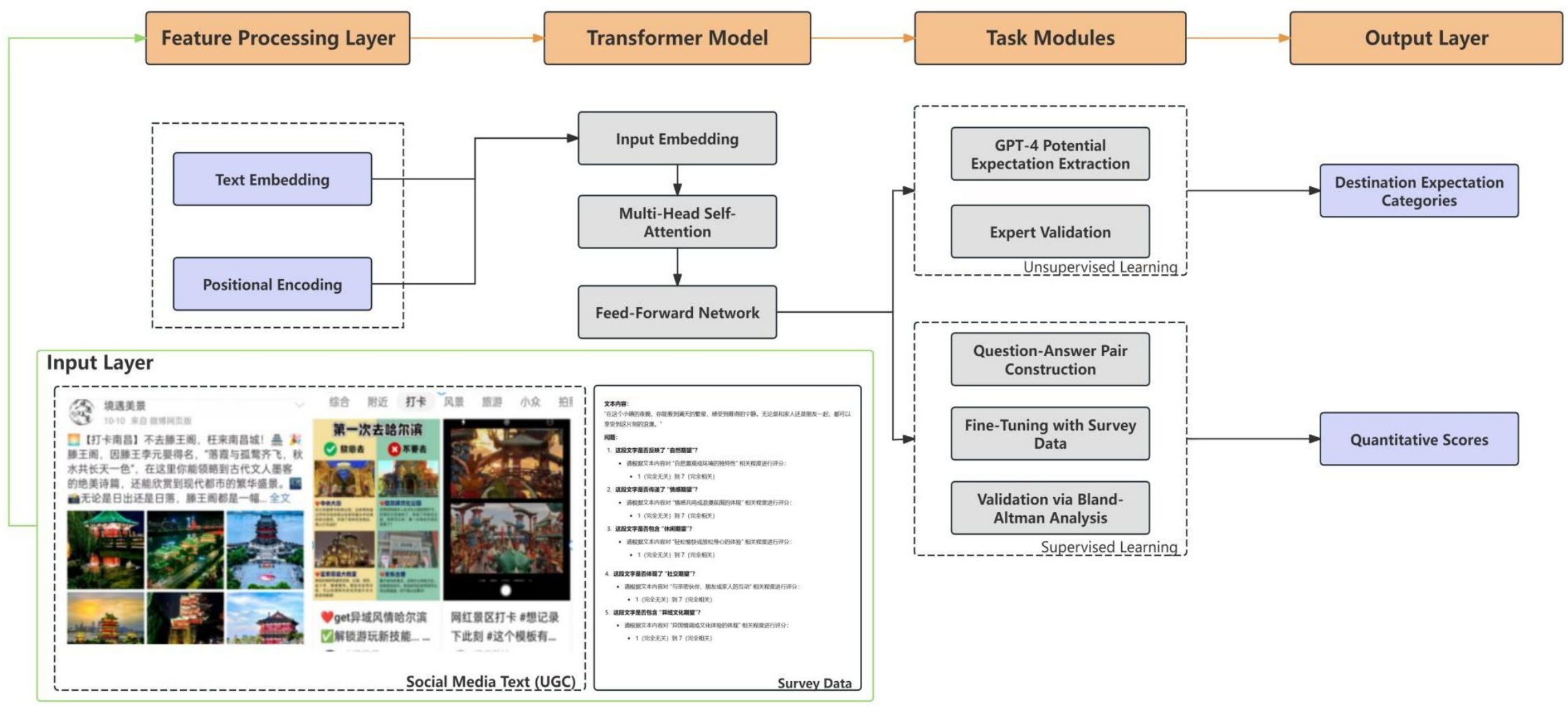

**Figure 1 Caption：** Model Architecture for Destination Expectation Analysis Using GPT-4

**Figure 1 Alt Text：** A flowchart depicting a pipeline where social media text is processed through a Transformer model, extracting potential expectations and validating them, leading to categorized outputs and quantitative scores.

*3.2.1 Unsupervised Learning: Analyzing Tourists' Latent Expectations*

In the unsupervised learning phase, we used the GPT-4 API to extract latent destination expectations mentioned by tourists in social media texts. By designing specific prompts, we were able to guide the model to generate content related to destination expectations. Each text was paired with a question, ultimately guiding the model to respond to queries such as, "Please analyze the expectations for destination X contained in this text." GPT-4 then generated an answer describing the latent expectations.

Each input social media text generated a list of potential expectations, which were automatically derived by the model based on contextual information. The processing formula for input text is as follows:

$$\hat{P} = f(\mathrm{Text}_{\mathrm{input}}; \theta_{\mathrm{GPT-4}})$$

Where $\hat{P}$ represents the set of latent destination expectations extracted from the input text, $\mathrm{Text}_{\mathrm{input}}$ is the preprocessed social media text, and $\theta_{\mathrm{GPT-4}}$ is the fixed function applied by the GPT-4 model.

To further refine the results, we conducted statistical analysis and filtering of all potential expectations in each sample. We then ranked the extracted expectations based on their word count proportion in the sample, selecting the most frequent destination expectations as the core of the study. This process can be expressed as:

$$P_{\mathrm{selected}} = \mathrm{Top} - k(\{\mathrm{Freq}(P_1), \mathrm{Freq}(P_2), \ldots, \mathrm{Freq}(P_n)\})$$

Where $P_{\mathrm{selected}}$ is the selected set of destination expectations, and $k$ is the number of high-frequency expectations chosen for the study.

The unsupervised paradigm capitalizes on GPT-4's emergent capability to derive contextual associations from pretraining corpora, obviating manual annotation through self-supervised pattern recognition. However, recognizing the epistemological uncertainty inherent in unsupervised discovery, we implemented a tripartite validation framework: 1) Expert panel review conducting stratified sampling of 10% model outputs; 2) Semantic consistency checks using BERT-based similarity scoring; 3) Temporal stability analysis across data collection waves. This hybrid verification approach aligns with recommended practices for validating LLM-derived constructs in social science research, ensuring the discovered expectations maintain face validity and theoretical coherence.

*3.2.2 Supervised Learning: Quantifying Tourists' Expectation Scores*

The supervised phase implemented parameter-efficient fine-tuning (PEFT) through Low-Rank Adaptation (LoRA), modifying only 0.3% of GPT-4's pre-trained parameters to preserve base model capabilities while adapting to expectation quantification tasks. Training data comprised <text, score, rationale> triples formatted as instruction-response pairs: "Given the social media text: [text], assign intensity scores (1-7 scale) for [expectation category] with explanatory rationale." This dual-task formulation enabled simultaneous regression (score prediction) and semantic alignment (rationale generation) through multi-objective optimization.↳

The architecture's self-attention mechanism was maintained in its original configuration to leverage GPT-4's native contextual understanding. Each input sequence concatenated: 1) Task instruction; 2) Raw social media text; 3) Expectation category identifier. Positional encoding preserved temporal relationships between textual elements, while layer normalization stabilized gradient flows during backpropagation.

The fine-tuning process employed cosine learning rate scheduling with 500 warmup steps, maintaining batch consistency across hardware accelerators through gradient checkpointing. Validation checks monitored loss convergence across three dimensions: score prediction MSE, rationale semantic similarity, and attention pattern consistency with base model baselines.

GPT-4 is a large-scale pre-trained language model based on the Transformer architecture, with robust text generation and comprehension capabilities. Its core architecture consists of self-attention layers and feed-forward networks. The self-attention layers capture the relationships between words in the text, allowing the model to consider contextual information when generating outputs. During fine-tuning, the model leverages its pre-trained language understanding capabilities to predict scores through API calls. The scores generated by the model can be represented using the following formula:

$$\hat{y} = f(\text{Text}_{\text{input}}; \theta)$$

Where $\hat{y}$ represents the expectation score generated by the model, $\text{Text}_{\text{input}}$ is the input social media text, and θ denotes the model's parameters.

To optimize the generated expectation scores, we calculate the difference between the predicted scores $\hat{y}$ and the true scores $y$. For this purpose, we use Mean Squared Error (MSE) as the loss function, defined as follows:

$$L(\theta) = \frac{1}{n}\sum_{i=1}^{n}(\hat{y}_i - y_i)^2$$

Where $\hat{y}_i$ is the predicted score for the i-th data point, $y_i$ is the corresponding true score, and $n$ represents the total number of samples.

Based on the loss function, the model uses the backpropagation algorithm to adjust its parameters, minimizing the loss through gradient descent. The parameter update rule is defined as follows:

$$\theta_{new} = \theta_{old} - \alpha \nabla_{\theta} L(\theta)$$

Where $\alpha$ represents the learning rate, $\nabla_{\theta} L(\theta)$ is the gradient of the loss function with respect to the model parameters, $\theta_{new}$ is the updated model parameter set, and $\theta_{old}$ is the current model parameter set.

Through this fine-tuning process, the model iteratively optimizes its parameters, resulting in more accurate expectation scores. After fine-tuning, the expectation scores generated by the model can be expressed as:

$$\hat{y}_{supervised} = f(Text_{input}; \theta_{new})$$

Where $\hat{y}_{supervised}$ represents the expectation score generated by the fine-tuned model, and $\theta_{new}$ denotes the updated model parameters after fine-tuning.

### *3.3 Validation and Working Principle of the Model*

To ensure the accuracy of the GPT-4 model in generating destination expectation scores, this study combined expert validation with automated verification.

In the unsupervised learning phase, where labeled data was not available, the destination expectations generated by the model were manually validated by an expert team. Composed of experienced industry professionals, the team reviewed and sampled the model's outputs to evaluate their alignment with actual tourist needs. This process ensured the accuracy, relevance, and practicality of the generated expectations.

In the supervised learning phase, the expert team not only examined the model-generated expectation scores but also compared them with UGC. Team members independently rated the UGC and compared their evaluations with the model's outputs to assess its precision in quantifying tourist expectations. This dual approach ensured that the model's outputs were consistent with real-world tourist demands, providing a robust and comprehensive validation process.

To further validate the accuracy of the model, we employed the Bland-Altman method to evaluate the consistency between the expectation scores generated by the model and the actual questionnaire scores. The Bland-Altman method assesses the differences between two sets of scores and determines their limits of agreement by calculating the mean difference and standard deviation. The formula is as follows:

$$\mu_D = \frac{1}{n}\sum_{i=1}^{n} D_i$$

Where $D_i$ represents the difference between the i-th pair of scores, and $\mu_D$ is the mean of the differences.

The formula for calculating the limits of agreement is as follows:

$$\text{Limits of Agreement} = \mu_D \pm 1.96 \cdot \sigma_D$$

Where 1.96 represents the commonly used coefficient for the 95% confidence interval, indicating that the score differences should fall within this range. If most differences lie within the limits of agreement, it suggests a high level of consistency between the model-predicted expectation scores and the actual scores. Conversely, if the differences fall outside this interval, it indicates significant deviations, suggesting that the model requires further optimization.

## 4. Results

### *4.1 Analysis of Tourism Expectations in User-Generated Content*

The unsupervised analysis revealed that GPT-4's latent space effectively captured nuanced tourism expectations embedded within pandemic-era discourse. A stratified sample of 3,000 social media posts underwent dual validation through expert consensus coding and semantic coherence evaluation, demonstrating the model's capacity to disentangle complex expectation constructs from vernacular expressions. Tourism professionals emphasized the model's proficiency in identifying context-dependent expectations, particularly noting its ability to differentiate between culturally-bound interpretations of "exotic experiences" across regional dialects—a task that challenges traditional NLP approaches.

The validation framework employed a 7-point relevance scale to assess expectation-text alignment, with ratings serving as qualitative indicators of model interpretability rather than quantitative metrics. This approach surfaced critical insights into LLM's reasoning patterns: model-generated justifications frequently referenced subtle linguistic cues such as metaphorical descriptions of natural landscapes ("the mountains whispered tranquility") as indicators of escapism expectations, and social media-specific neologisms ("staycation blues") signaling emerging leisure demands.

Interdisciplinary evaluation highlighted two key strengths of the methodology: First, the integration of expert validation mitigated the epistemological uncertainty inherent in unsupervised expectation discovery, ensuring construct validity through iterative reconciliation of machine outputs with tourism psychology frameworks. Second, the model's chain-of-thought explanations provided audit trails for expectation attribution, enabling researchers to trace how specific lexical patterns (e.g., co-occurrence of temporal markers and emotional adjectives) informed final expectation categorizations.Ultimately, we obtained the tourism expectation information corresponding to each sample (Table 1).

**Table 1.** Example of Extracted Tourism Expectations from User-Generated Content

| Expectation Category | Subcategory | Word Count | Content (%) | Score (1-7) | Expression | Inference |
|---|---|---|---|---|---|---|
| Emotional Expectations | Romantic Experience | 55 | 10% | 7 | - With loved one<br>Romantic<br>- desert journey | - “Experience the desert with someone special”<br>- “A sunrise that |
| | | | | | | turns the sky red” |
| | Memorable Experience | 15 | 3% | 5 | - Unforgettable experience | - “I guarantee it will be unforgettable” |
| | Relaxation & Comfort | 20 | 4% | 4 | - Night market strolls<br>- Comfortable clothes | - “Wandering the night market in comfort” |
| Natural Expectations | Unique Natural Scenery | 80 | 15% | 7 | - Deserts, lakes, parks<br>- Beautiful landscapes | - “Tengger Desert, one of the six most beautiful deserts in China” |
| | Natural Phenomena | 25 | 5% | 6 | - Starry nights<br>- Sunrises | - “A sky full of stars”<br>- “Watching the sunrise” |
| Exotic Culture Expect. | Cultural Experience | 50 | 9% | 6 | - Southeast Asian<br>- architecture<br>- Foreign culture | - “Golden pagodas, Southeast<br>- Asian-style buildings” |
| | Domestic Exotic Exper. | 45 | 8% | 7 | - Foreign beauty within China | - “Experience Morocco-like landscapes in China” |
| Budget Expectations | Low-Cost Travel | 70 | 13% | 6 | - Affordable destinations<br>- Budget travel | - “Only 1000 per person for these places” |
| | Value for Money | 50 | 9% | 5 | - Cost-effective travel<br>- High value for | - “Feel like you’re in Thailand for 300” |
| | | | | | price | |
| Uniqueness Expectations | Hidden Gems | 40 | 7% | 7 | - Lesser-known spots<br>- Hidden gems | - “A hidden treasure, rarely mentioned but worth it” |
| | … | … | … | … | … | … |

By analyzing the expectation content generated by the model, we found significant thematic similarities among many expectations. To further refine the analysis, similar tourism expectations were consolidated to ensure that the most representative categories were extracted. The consolidated expectations were ranked based on their proportion, resulting in the identified tourism expectation categories (Figure 2).

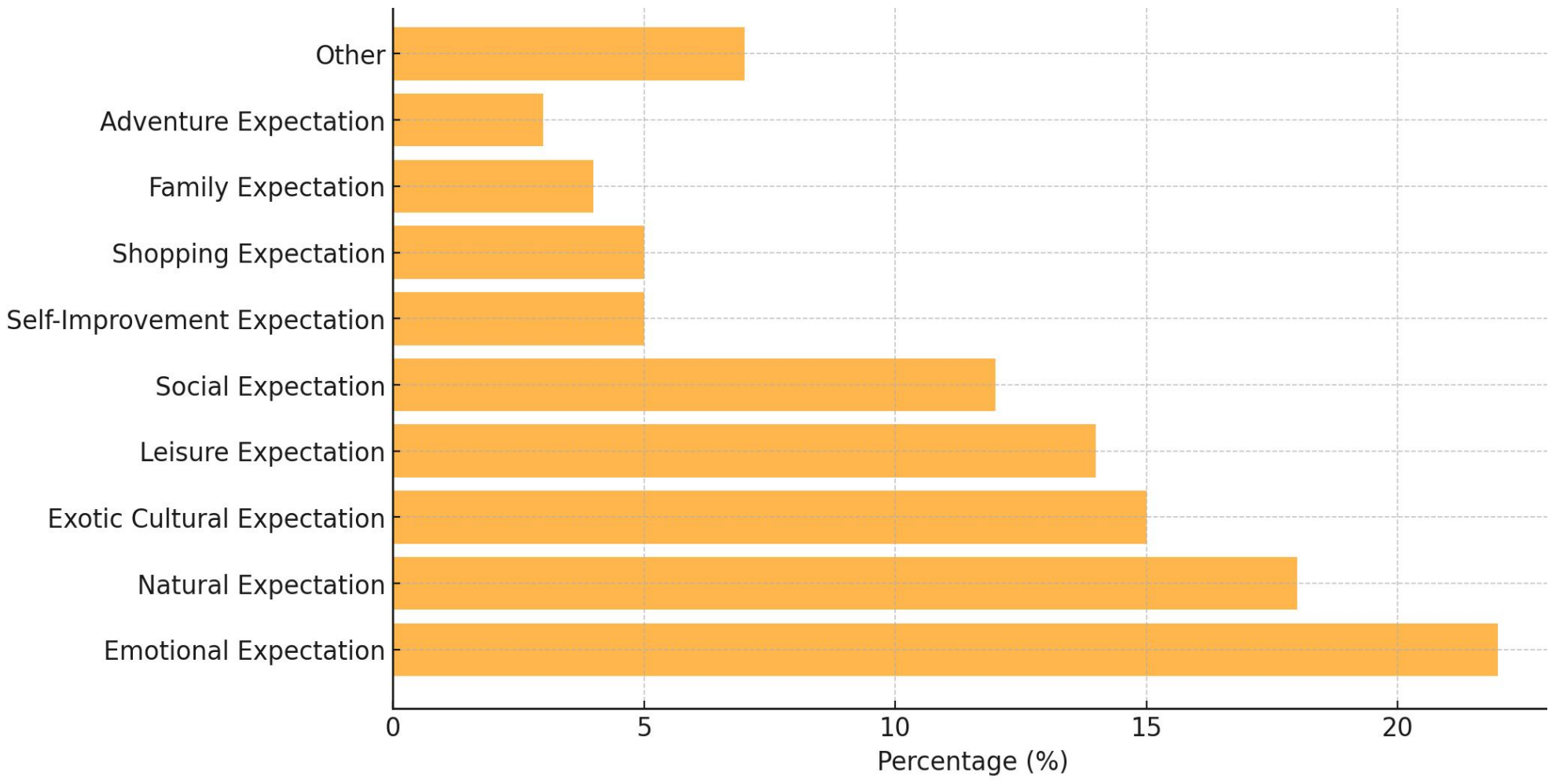

**Figure 2 Caption:** Travel Expectation Category Proportions

**Figure 2 Alt Text:** A bar chart showing different expectation types, with percentages on the x-axis and categories on the y-axis

After completing the consolidation and ranking, we selected the top five tourism expectations - Emotional Expectations, Natural Expectations, Exotic Cultural Expectations, Relaxation Expectations, and Social Expectations - as the focus of this study (Table 2). These categories represent the core needs of tourists during travel, with their proportions and relevance standing out across all samples.

**Table 2.** Descriptions of Selected Tourism Expectations

| Tourism Expectation | Description |
|---|---|
| Emotional Expectation | Focused on romantic or memorable travel experiences shared with loved ones, creating emotional resonance. |
| Natural Expectation | Emphasizes appreciation of unique natural landscapes, such as deserts, lakes, and mountains. |
| Exotic Cultural Expectation | Highlights exploration of cultural atmospheres, including regional architecture and traditional activities. |
| Leisure Expectation | Prioritizes relaxation and enjoyable travel activities, including culinary experiences and comfortable accommodations. |
| Social Expectation | Reflects the importance of traveling with companions and sharing experiences through social media. |

### *4.2 Model Fine-Tuning and Tourism Expectation Scoring*

The supervised learning framework utilized 7,135 user-generated content entries (70% of total samples) for expectation quantification modeling. Adhering to few-shot learning principles, the training subset was strategically limited to 100 representative samples (1.4% of total data), with 1,425 entries (28.5%) allocated for validation—maintaining a balance between model adaptability and evaluation rigor.

The validation protocol engaged 1,500 distributed evaluators through crowdsourcing platforms, achieving multi-perspective assessment with 15 independent ratings per sample. This design ensured robust estimation of expectation intensity while capturing subjective variance in tourism perception. Questionnaire items focused on three evaluative dimensions: contextual relevance of identified expectations, accuracy of emotional attribution, and consensus level regarding expectation categorization.↳

Parameter optimization was implemented through feedback integration, where original model outputs were systematically replaced with human-annotated scores to construct supervised signal pairs. The fine-tuning process transformed expectation identification tasks into regression problems, enabling the model to internalize the nuanced relationship between linguistic patterns and expectation intensity levels.

Validation employed the Bland-Altman method to assess agreement between machine-generated scores and human evaluations. Given the multi-rater design, final scores were aggregated through mean value synthesis prior to comparison. The graphical analysis demonstrated sufficient congruence between computational predictions and human judgments, with majority residual values residing within predefined confidence boundaries (Figure 3). This concordance confirms the model's capacity to emulate expert-level expectation quantification in tourism UGC analysis.

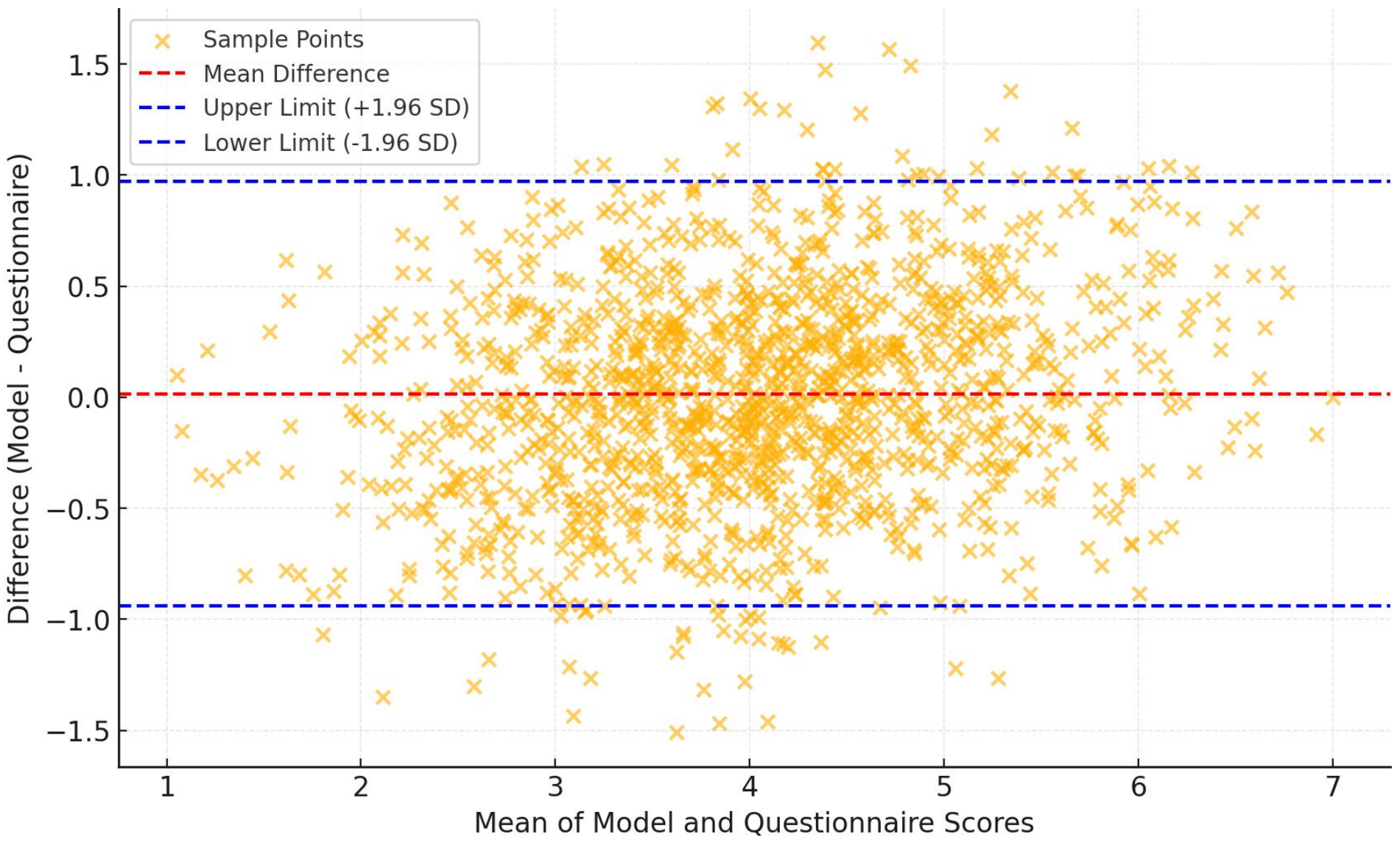

**Figure 3 Caption：** Bland-Altman Plot for Model-Generated and Questionnaire Scores

**Figure 3 Alt Text：** Bland-Altman plot showing mean differences between model and questionnaire scores, with limits of agreement and sample points.

After fine-tuning the model and validating its accuracy, we obtained scores for the five primary tourism expectations across all user-generated content. These scores represent the intensity of each expectation within individual posts, highlighting the most prominent tourism expectations in each piece of user-generated content.

### *4.3 Relationship Between Tourism Expectations and Likes: Experimental Analysis*

The analysis employed iterative computation to enhance measurement reliability, with each expectation score undergoing ten independent model evaluations followed by mean value aggregation. To address potential distortion from exceptionally high engagement metrics, like counts were proportionally scaled by a factor of 100, preserving relative differences while mitigating variance inflation.

Initial exploration through multivariate linear regression modeled like counts as a function of expectation intensity scores (Table 3). The results indicated marginal positive associations between specific expectation dimensions and user engagement, though with limited practical significance evidenced by small regression coefficients.The overall model demonstrated constrained explanatory capacity, suggesting the necessity for alternative analytical frameworks to fully capture the complex interplay between psychological expectations and social media interactivity.

**Table 3.** Linear Regression Results for Tourism Expectations and Likes

| | Variable | Coefficient | Std Error | t-Value | P-Value |
|---|---|---|---|---|---|
| const | Intercept | 258.14 | 18.63 | 13.86 | ＜0.05 |
| Emotional Expectation | Emotional Expectation | -1.49 | 2.03 | -0.73 | 0.46 |
| Natural Expectation | Natural Expectation | 0.34 | 2.01 | 0.17 | 0.86 |
| Exotic Cultural Expectation | Exotic Cultural Expectation | 3.52 | 2.05 | 1.72 | 0.09 |
| Leisure Expectation | Leisure Expectation | 7.08 | 2.01 | 3.52 | ＜0.05 |
| Social Expectation | Social Expectation | 5.61 | 2.05 | 2.74 | ＜0.05 |

The random forest model was employed to further analyze the complex relationship between tourism expectations and the number of likes. This model effectively captures nonlinear features and identifies the contribution of each tourism expectation to the like counts. Using the model, we determined the importance weights and contribution of each tourism expectation to user interactions.

To interpret the random forest model's results, we utilized Partial Dependence Plots (PDPs) and Shapley Additive Explanations (SHAP) analysis. PDPs visually illustrate the influence of specific features on the model's predictions, while SHAP analysis accurately explains the impact of each tourism expectation on the predicted like counts. Together, these methods revealed how different expectations affect user engagement and provided detailed insights into the influence of each expectation on user attention.

The analysis results of the random forest model (Figure 4) indicate that Exotic Cultural Expectation and Natural Expectation have the highest feature contributions, both at 0.207. Emotional Expectation follows with a feature contribution of 0.204, while Social Expectation contributes 0.197. Leisure Expectation has the lowest feature contribution at 0.185.

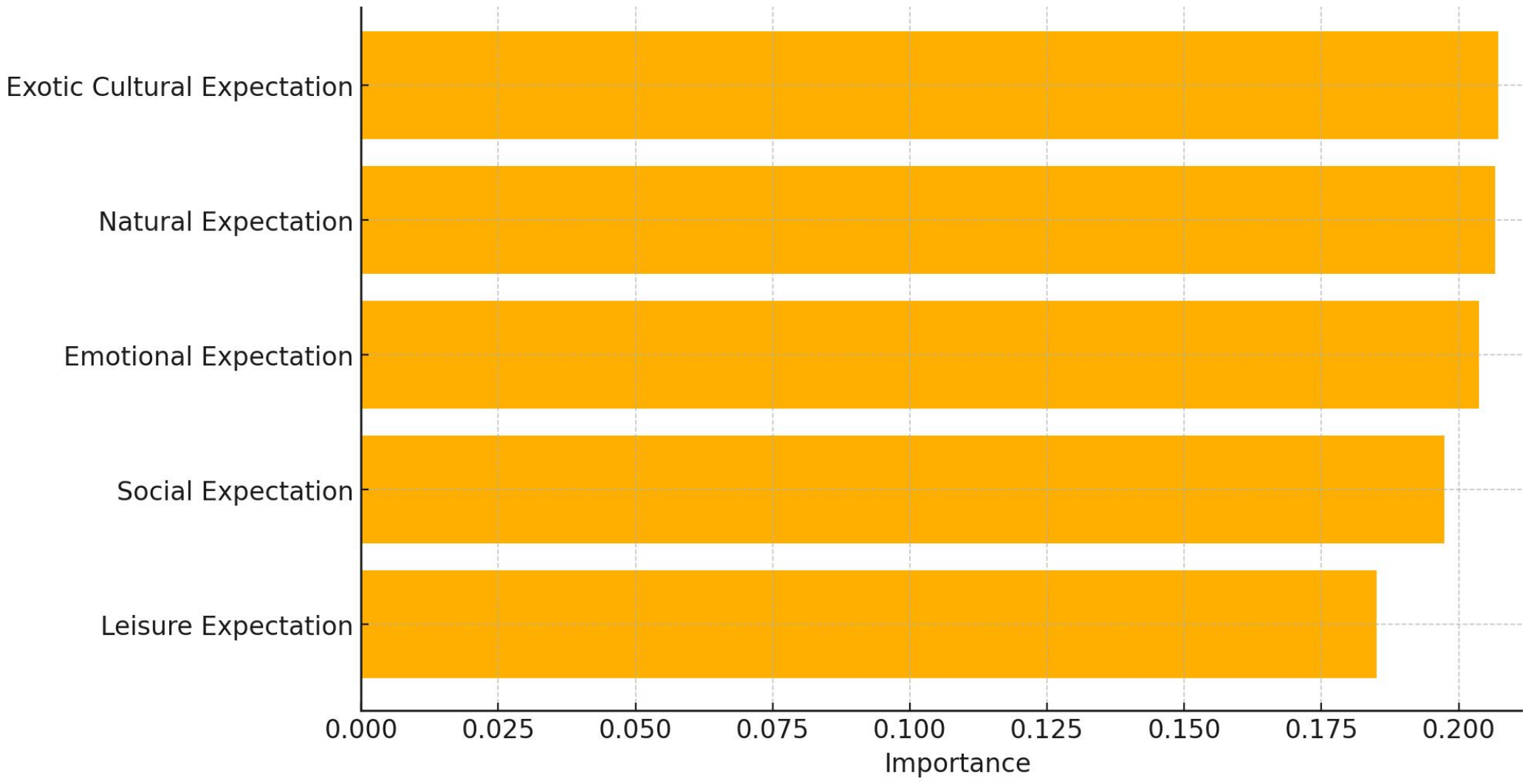

**Figure 4 Caption：** Feature Contributions of Tourism Expectations in the Random Forest Model

**Figure 4 Alt Text：** Bar chart displaying the importance of different expectations: Exotic Cultural, Natural, Emotional, Social, and Leisure.

The partial dependence plots (Figure 5) illustrate the influence trends of various features on the target variable. The curve for Emotional Expectation exhibits significant fluctuations, with the target variable showing notable changes when the feature values approach 0 and the positive range. The curve for Natural Expectation is relatively stable, displaying a steady upward trend as the feature values increase. The influence curve for Exotic Cultural Expectation demonstrates a clear positive growth in the positive feature value range, accompanied by moderate fluctuations. Leisure Expectation exhibits pronounced nonlinear characteristics, with substantial variations in the negative feature value range and higher target variable values in the positive range. Finally, the curve for Social Expectation shows a consistent upward trend, with the target variable increasing steadily as feature values rise.

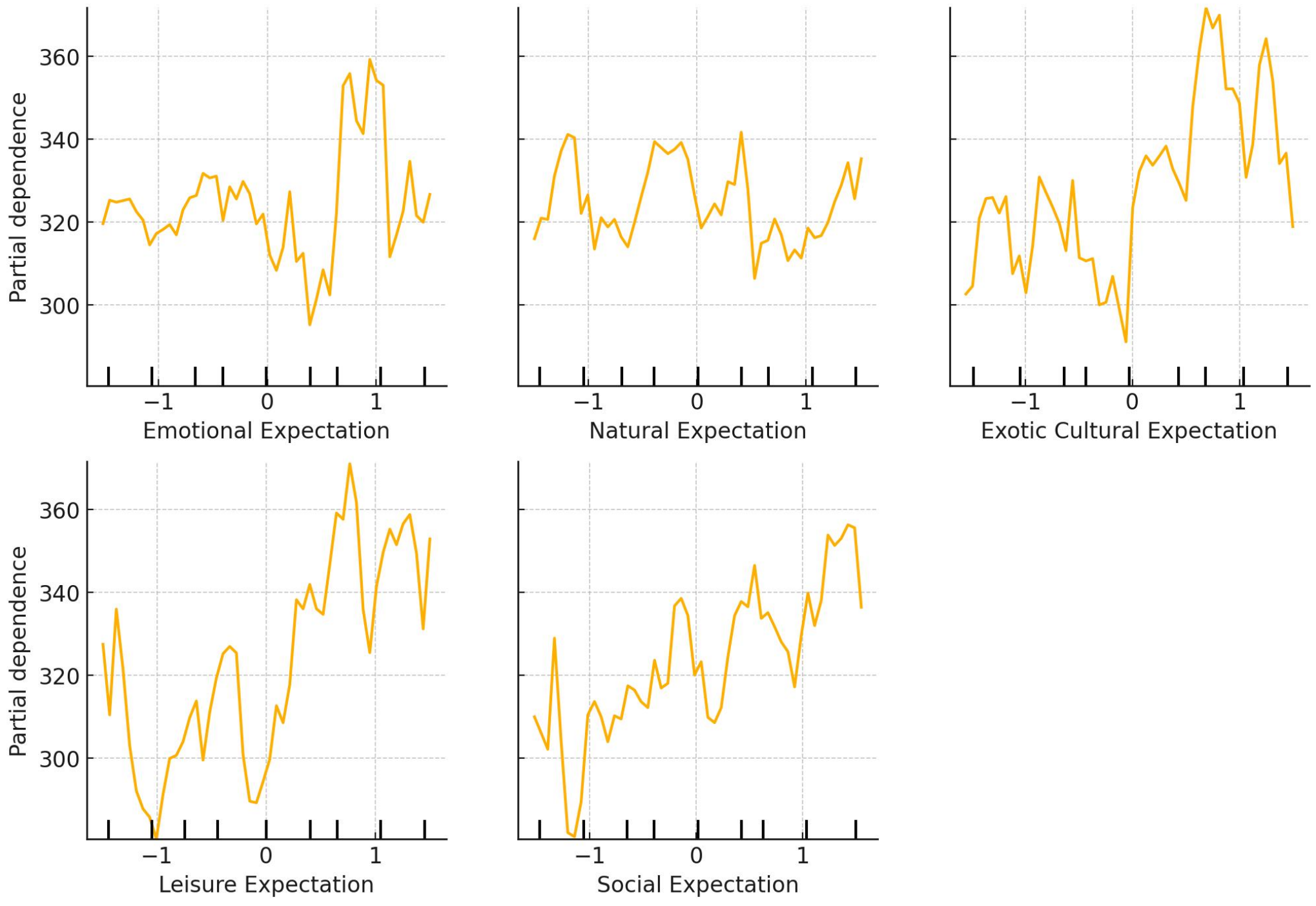

**Figure 5 Caption：** Partial Dependence Plots of Tourism Expectations on the Target Variable

**Figure 5 Alt Text：** Partial dependence plots showing the effect of five expectations: Emotional, Natural, Exotic Cultural, Leisure, and Social on the model.

The SHAP summary plot (Figure 6) provides an overview of the local contributions of various features across different samples. The upper line chart illustrates the overall trend of model predictions (f(x)), while the lower heatmap reveals the direction and intensity of each feature’s contribution to the predictions.From the heatmap, Leisure Expectation and Social Expectation show the most prominent red regions, indicating strong positive contributions to the target variable in the majority of samples. In contrast, Emotional Expectation and Exotic Cultural Expectation exhibit significant individual variability, with both positive and negative impacts. Natural Expectation displays a more neutral contribution, as reflected by the balanced distribution of red and blue regions, suggesting a relatively stable influence across samples.

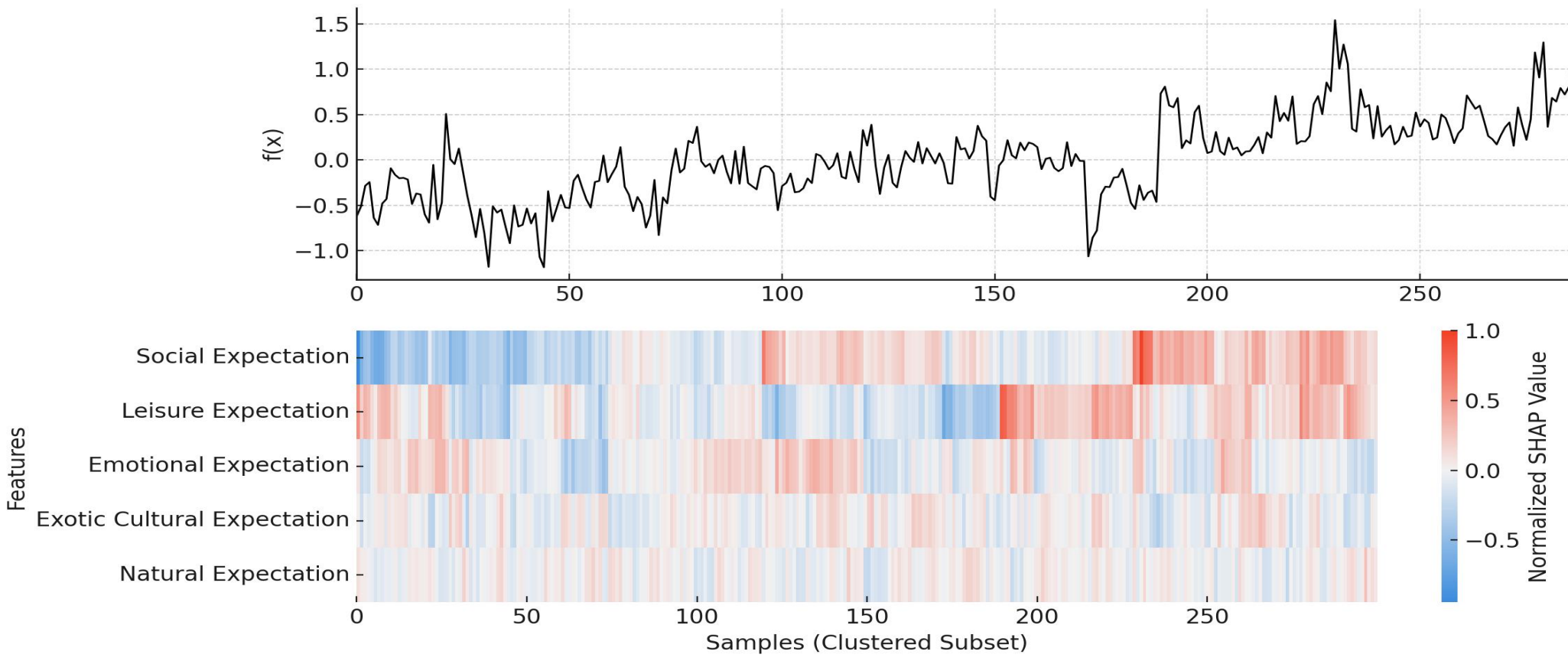

**Figure 6 Caption：** SHAP Summary Plot of Feature Contributions to the Target Variable

**Figure 6 Alt Text：** SHAP summary plot showing individual sample feature importance for Social, Leisure, Emotional, Exotic Cultural, and Natural Expectations.

Although Natural Expectation, Emotional Expectation, and Exotic Cultural Expectation demonstrated higher contributions in feature importance analysis, their positive impact on "likes" appears limited based on SHAP values and PDP plots. These features exhibit significant variability and uncertainty in their influence across different ranges or samples.

In contrast, while Leisure Expectation and Social Expectation show lower overall feature importance, they display more prominent positive contributions (as indicated by the larger red regions in the SHAP heatmap). This suggests that these features have a more stable and practical influence on increasing "likes."

**5. Discussion**

The empirical results reveal a critical divergence between feature salience and behavioral impact in tourism expectation dynamics. Natural and Emotional Expectations, despite their structural prominence in prediction models, exhibit attenuated effect magnitudes on digital engagement metrics. This paradox suggests a maturation phase in destination perception formation, where foundational tourism dimensions achieve cognitive entrenchment at the expense of behavioral stimulation - a phenomenon aligning with the expectation saturation threshold theory in experience economy research.

Exotic Cultural Expectation manifests a dual-nature influence characteristic of trend-mediated constructs. While demonstrating positive valence in both global feature ranking and marginal effect analysis, its SHAP value distribution reveals susceptibility to temporal volatility and demographic specificity. This behavioral pattern corresponds to the conceptual framework of transitory destination image formation, where exogenous cultural elements achieve episodic salience through mediated content virality rather than enduring cognitive schema evolution.

The countervailing dominance of Leisure and Social Expectations unveils a fundamental restructuring of tourism engagement paradigms. Their robust effect consistency across analytical dimensions establishes these dimensions as primary catalysts in the contemporary digital engagement ecosystem. This empirical evidence substantiates the compensatory gratification hypothesis, suggesting that social media platforms have transitioned from mere information channels to constitutive spaces for hedonic fulfillment and parasocial interaction - a transformation accelerated by pandemic-induced behavioral recalibration.

The COVID-19 context emerges as a critical moderator in expectation-behavior decoupling phenomena. Traditional core expectations (Natural/Emotional) maintained latent cognitive representation but suffered behavioral actualization deficits due to mobility constraints. Conversely, Exotic Cultural content fulfilled symbolic compensation functions through imaginative transportation, while Leisure/Social Expectations enabled immediate need gratification through digital proxemics - a bifurcation that exemplifies the adaptive resilience framework in crisis tourism.

Methodologically, the LLM-driven expectation quantification paradigm introduces three substantive advancements: (1) semantic topology reconstruction through neural linguistic mapping, overcoming lexical fragmentation in traditional content analysis; (2) dynamic expectation intensity calibration via transfer learning with psychometric data; (3) real-time behavioral prediction capability through deep feature fusion architecture. This tripartite innovation establishes a new protocol for tourism expectation operationalization in the era of generative AI, particularly salient given the exponential growth of multimodal UGC.

The study's theoretical contribution resides in its deconstruction of the expectation-engagement nexus through a computational hermeneutics lens. By empirically validating the pandemic-induced transition from destination-centric to experience-centric digital tourism paradigms, we extend the applicability of the uses and gratifications theory to hybrid physical-digital consumption contexts. Furthermore, the identified expectation hierarchy provides a diagnostic framework for destination marketing organizations to optimize content strategies in platform-mediated environments.

## 6. Conclusions

This research establishes a novel computational framework for tourism expectation analytics through the integration of large language models (LLMs) and psychometric validation. The proposed methodology demonstrates three pivotal advancements: First, it enables automated extraction of latent tourism expectations from unstructured social media discourse through neural semantic decomposition; Second, it operationalizes abstract psychological constructs into quantifiable metrics via human-AI hybrid intelligence; Third, it reveals platform-mediated behavioral shifts where digital-native expectations (Leisure/Social) supersede traditional tourism drivers in engagement generation.

The empirical findings illuminate a fundamental reconfiguration of tourist psychology in algorithmically-curated environments. While Natural and Emotional Expectations retain structural prominence in feature importance metrics, their diminished predictive power for engagement outcomes underscores the emergence of platform-specific behavioral paradigms. The dichotomy between trending cultural content's transient appeal and leisure/social expectations' sustained influence exposes the tension between algorithmic amplification and authentic user demand in digital tourism ecosystems.

Methodologically, the study contributes a reproducible pipeline for tourism text analysis, combining LLMs' contextual awareness with survey-based validation to bridge computational outputs and psychological constructs. The demonstrated capability to process vernacular expressions at scale while maintaining theoretical coherence addresses longstanding challenges in tourism social media research.

Current limitations regarding data platform dependency and prompt sensitivity delineate clear pathways for future inquiry. Subsequent research should prioritize 1) Cross-cultural validation of expectation hierarchies through multilingual analysis, 2) Multimodal integration of visual-textual data streams, and 3) Longitudinal tracking of expectation evolution in post-pandemic contexts. This methodological trajectory positions LLMs as indispensable tools for decoding digital-era tourism psychology, offering transformative potential for destination experience management and personalized content optimization.

**Disclosure statement**

No potential conflict of interest was reported by the author(s).

**Funding**

This research was funded by National Key Research and Development Program of China (Grant No. 2022YFD1600500), "Research and Application Demonstration of Key Technologies for the Cold-Region Specialty Fruit Tree Industry."